\documentclass[pmlr,twocolumn,10pt]{jmlr} 

\mlhtrack{proceedings}

\newif\iffinal
\finaltrue  

\iffinal
\ifmlhneedspmlr
\jmlrvolume{346}
\jmlryear{2026}
\fi
\ifmlhfindings \jmlrproceedings{}{ML4H 2026 - Findings Track}\fi
\ifmlhdemo     \jmlrproceedings{}{ML4H 2026 - Demo Track}\fi
\jmlrworkshop{Machine Learning for Health (ML4H) 2026}
\else
\jmlrproceedings{}{Submitted to ML4H 2026: \mlhtrackname}
\jmlrworkshop{Machine Learning for Health (ML4H) 2026}
\fi

\usepackage{booktabs}
\usepackage{siunitx}

\usepackage[switch]{lineno}

\usepackage{tikz}
\usetikzlibrary{positioning,arrows.meta}
\usepackage{algorithm, algpseudocode}

\theorembodyfont{\upshape}
\theoremheaderfont{\scshape}
\theorempostheader{:}
\theoremsep{\newline}

\title[Uplift Modelling of a Nationwide Home Blood Pressure Monitoring Programme]{Who Benefits from Telemonitoring? Uplift Modelling of a Nationwide Home Blood Pressure Monitoring Programme}

\author{
	  \Name{Connie Kou} \Email{connie.kou@mohh.com.sg} \\
	  \addr MOH Office for Healthcare Transformation, Singapore
}

\begin{document}
	
	\maketitle
	
	\ifmlhdemo\else
	\begin{abstract}
		Home blood pressure (BP) telemonitoring programmes are being deployed at national scale, yet enrolment resources are finite and the benefit of telemonitoring is unlikely to be uniform across patients. We study the Primary Technology-Enhanced Care for Hypertension (PTEC-HT) programme, a nationwide telemonitoring programme for hypertensive patients in Singapore. Using retrospective data on 16,694 patients registered between August 2020 and May 2024, we first estimate the average treatment effect of the programme on 6-month BP control against a propensity-score-matched non-participant cohort, finding an absolute improvement of 5 percentage points in the rate of BP control (74\% vs.\ 69\%, $p = 3\times10^{-12}$), rising to 12 percentage points (63\% vs.\ 50\%, $p = 1\times10^{-15}$) among patients with uncontrolled BP at baseline. We then move beyond the average effect and apply uplift modelling to estimate individual treatment effects. Across five uplift learners evaluated with 5-fold cross-validation and Qini curves, an uplift random forest performed best on the full cohort and an S-learner on the uncontrolled-baseline subcohort, although performance bands overlapped across learners. On held-out patients, realised uplift was largest in the top predicted-uplift groups and declined across the ranking. These patients had higher baseline BP grade, more recent hypertension diagnosis and lower medication and diagnosis burden, a profile we offer as hypothesis-generating for targeted enrolment. A handcrafted rule built from these features was competitive with the learned models, outperforming all of them on the uncontrolled-baseline cohort, suggesting that uplift models are useful for discovering heterogeneity that can be distilled into simple enrolment criteria for directing limited care team capacity.
	\end{abstract}
	\fi
	
	\ifmlhneedsstatements
	\paragraph*{Data and Code Availability}
	This study uses deidentified telemonitoring programme data and electronic health records from PTEC-HT and national primary care databases in Singapore, comprising demographics, hypertension diagnoses, antihypertensive medication prescriptions and clinic BP measurements. Owing to patient privacy requirements and national data governance policies, the data and code cannot be made publicly available. Pseudocode for the matching and uplift pipeline is provided in \appendixref{apd:pseudocode}.
	
	\paragraph*{Institutional Review Board (IRB)}
	This research does not require IRB approval. It is a retrospective secondary analysis of de-identified data collected during routine care under the PTEC-HT programme, with no intervention on or interaction with patients and no use of individually identifiable information.
	\fi
	
	\section{Introduction}
	\label{sec:intro}
	
	Hypertension is a leading modifiable risk factor for cardiovascular disease and its burden is growing rapidly in ageing populations. In Singapore, the Primary Technology-Enhanced Care (PTEC) programme was launched in 2020 to help patients manage chronic conditions from home \citep{wang2025economic}. Its first module, Primary Technology-Enhanced Care for Hypertension (PTEC-HT), provides hypertensive patients with a Bluetooth-enabled BP monitor and a companion phone app for at least weekly home measurements. Readings are transmitted automatically to a vital signs monitoring dashboard used by polyclinic care teams who receive automated alerts on abnormal or missed readings, conduct teleconsultations and adjust medications without requiring clinic visits.
	
	Telemonitoring interventions for hypertension have shown positive but heterogeneous effects in trials and meta-analyses \citep{tucker2017}. Two questions matter for a programme operating at national scale. First, \emph{does the programme improve BP control on average}, relative to comparable patients receiving usual care? Second, and operationally more important, \emph{for whom does it work}? Care-team capacity, devices and onboarding effort are finite. If the benefit is concentrated in an identifiable subgroup, recruitment can be prioritised accordingly.
	
	The second question is one of heterogeneous, individual-level treatment effects rather than a single average effect. Uplift modelling \citep{radcliffe2007, gutierrez2017} addresses exactly this: it estimates for each individual, the difference between their outcome probability under treatment and under no treatment,
	\begin{equation}\label{eq:uplift}
		\tau(x) = P(Y{=}1 \mid T{=}1, X{=}x) - P(Y{=}1 \mid T{=}0, X{=}x),
	\end{equation}
	where $Y$ is BP control at follow-up, $T$ is programme participation, and $X$ are baseline covariates. In the standard uplift taxonomy, patients divide into \emph{persuadables} (positive uplift, controlled only if treated), \emph{sure things} (controlled regardless), \emph{lost causes} (uncontrolled regardless) and \emph{sleeping dogs} (negative uplift). Among these groups, only persuadables justify the marginal cost of intervention. Because individual counterfactuals are unobserved, we use `persuadable’ throughout to denote a patient with a positive estimated CATE given baseline covariates, not an identified individual-level type.
	
	\paragraph{Contributions}
	To our knowledge, this is among the first uplift-modelling analyses of a nationwide hypertension telemonitoring programme. Specifically, (i) we construct propensity-score-matched treated and control cohorts from national primary care data and estimate the average effect of PTEC-HT participation on 6-month BP control, (ii) we train and compare five uplift learners, selecting models by Qini curve analysis under 5-fold cross-validation, (iii) we characterise patients with higher predicted uplift, showing that realised uplift is larger in the top predicted-uplift groups, (iv) we distil the model features into a handcrafted rule that is competitive with the learned models and (v) we repeat the full pipeline on the clinically most relevant subcohort, patients with uncontrolled BP at baseline, since most enrolees are already controlled and hence have limited room to show improvement.

	\section{Related Work}
	\label{sec:related}
	
	\paragraph{BP Telemonitoring}
	Randomised trials and meta-analyses of home BP telemonitoring, particularly when combined with co-interventions such as clinician-led medication titration, report clinically meaningful reductions in systolic BP \citep{tucker2017}. The PTEC-HT programme itself was preceded by a quasi-experimental pilot trial in a single Singapore polyclinic showing improved BP control and cost-effectiveness relative to usual care \citep{teo2021ptecpilot}, and a subsequent real-world evaluation found the programme cost-saving and more effective in controlling BP status over 12 months \citep{wang2025economic}. Qualitative work alongside the pilot has documented implementation barriers and facilitators \citep{teo2023sociotechnical}, and trust relationships \citep{chew2023trust}. Most of these evaluations, however, report only average effects. Heterogeneity is typically explored through pre-specified subgroups rather than learned from data.
	
	\paragraph{Uplift Modelling and Heterogeneous Treatment Effects}
	Uplift modelling originated in marketing response modelling \citep{radcliffe2007} and is closely connected to the conditional average treatment effect (CATE) estimation literature in causal machine learning. Meta-learner strategies, S-, T-, X- and R-learners, reduce CATE estimation to standard supervised learning \citep{kunzel2019, nie2021}, while tree- and forest-based methods modify split criteria to directly target treatment effect heterogeneity \citep{rzepakowski2012, athey2019}. Model comparison in the absence of ground-truth individual effects commonly relies on the Qini curve and its area \citep{radcliffe2007}. Applications in healthcare remain comparatively rare, with prior work in treatment prioritisation \citep{li2024rrt} and clinical alerting \citep{biswas2018}.

	\section{Data and Cohort Construction}
	\label{sec:data}
	
	\subsection{Source Data and Cohorts}
	We identified 16,694 patients registered with PTEC-HT between August 2020 and May 2024. Because the outcome requires a clinic-measured BP during follow-up, the primary (\emph{full}) cohort comprises $8{,}563$
	PTEC-HT patients with at least one clinic BP measurement in the 6 months after enrolment. A secondary (\emph{uncontrolled BP}) cohort comprises $2{,}976$ patients whose average BP over the 1-year baseline window was uncontrolled and who likewise had a clinic BP in the following 6 months. \figureref{fig:flow} in \appendixref{apd:flow} traces how each cohort was derived.
	
	For each treated patient, the observation time is the PTEC-HT registration date and the outcome BP control is computed over the subsequent 6-month intervention period. Control candidates are drawn from the non-PTEC-HT hypertensive polyclinic population, with observation time assigned as a random date between August 2020 and May 2024 to mirror the enrolment distribution. The patient cohorts are split into train and test sets following an 80-20 split. The outcome is binary BP control, defined from the average of clinic BP measurements in the 6-month post-observation window, graded against standard clinic BP threshold of $<140/90$\,mmHg.

	\subsection{Propensity Score Matching} \label{sec:psm}
	In the absence of a randomised trial, we matched each PTEC-HT patient to a similar non-PTEC-HT patient by propensity score matching \citep{rosenbaum1983}. Propensity scores were estimated using logistic regression, with PTEC-HT participation regressed on: age, gender, years since first hypertension diagnosis, number of coded hypertension diagnoses (all-time), count of antihypertensive medication prescriptions (all-time), baseline BP grade over three time windows (past 3 months, 1 year and 5 years), household size, Charlson Comorbidity Index (CCI) and subsidy tier under the Community Health Assist Scheme (CHAS). Full definitions of each covariate are provided in \appendixref{apd:model_featdescription}. Demographic features (age, gender, household size) were available for all patients. Patients without a baseline clinic BP measurement in the relevant lookback window were assigned Normal grade BP. Patients without coded hypertension diagnoses or antihypertensive medication prescriptions in the records were assumed to have zero diagnoses and zero prescriptions respectively. No other imputation was applied to any covariate.
	
	Each PTEC-HT patient was matched 1:1 to the nearest-neighbour non-PTEC-HT candidate by propensity score without replacement and within a caliper of 0.01 standard deviations of the estimated propensity score. We estimate propensity scores and perform matching separately within each cohort. 85\% of eligible PTEC-HT patients in the full cohort and 69\% in the uncontrolled BP cohort were successfully matched within caliper, and unmatched patients are excluded from the corresponding analysis. Match quality was assessed by standardised mean differences (SMD) for all covariates before and after matching. After matching, all covariates achieved SMD below 0.1, indicating good balance 	(\appendixref{apd:all_smd}, \tableref{tab:smd_matched}). The train-test split and all 5-fold CV folds were formed at the matched-pair level, keeping each pair together.
	
	Unmatched patients differed systematically from matched patients (\appendixref{apd:matched_vs_unmatched}, \tableref{tab:matched-unmatched}). In both cohorts they were younger, more recently diagnosed and had higher baseline BP grades, consistent with lying in the high-propensity tail where few non-participants have comparable profiles.
	
	\section{Methods}
	\label{sec:methods}
	
	\subsection{Average Treatment Effect (ATE)}
	On the matched cohorts we compare the rate of BP control at the end of the 6-month intervention period between the PTEC-HT and control groups, testing the difference in proportions. Baseline rates are compared to verify that matching yields comparable starting points.
	
	\subsection{Uplift Models}	\label{sec:methods_upliftmodels}
	We estimate the individual uplift in \equationref{eq:uplift} with five learners: S-, T-, X- and R-learner meta-learners and an uplift random forest with treatment-effect-targeted split criteria \citep{kunzel2019, nie2021, rzepakowski2012}. Models are trained on the matched data using the same covariates as the propensity score model. The four meta-learners differ in how they use the treatment indicator. The S-learner includes it as a single input feature, the T-learner fits separate models per treatment arm, the X-learner extends this by imputing individual effects and re-weighting by propensity score, and the R-learner targets the effect directly via residual-on-residual regression. The uplift random forest instead modifies the tree-splitting criterion to directly maximise treatment-effect heterogeneity between leaves. All uplift models were implemented using the \texttt{causalml} Python library \citep{chen2020causalml}. Pseudocode for the matching and uplift pipeline is provided in \appendixref{apd:pseudocode} and full hyperparameters for both cohorts in \appendixref{apd:model_hyperparam}. Base learners and hyperparameters were selected separately for the full and uncontrolled BP cohorts via cross-validation, spanning logistic regression, random forest and L2-regularised linear/SGD classifiers.

	\subsection{Evaluation: Qini Curves}
	Since individual counterfactuals are unobserved, models are compared with Qini curves: patients are ranked by predicted uplift in descending order, and the cumulative gain in the targeted fraction (treated-group responders minus control-group responders, rescaled to equal group sizes) is computed as the targeted fraction grows. A better model concentrates positive uplift at the top of the ranking, yielding a larger area under the Qini curve (AUQC). We select models by AUQC on held-out folds under 5-fold cross-validation performed on the training set and final model comparison is illustrated via Qini curves on the test set.

	The Qini curve also has a direct operational reading: its right endpoint is the average effect of treating everyone and a rising segment identifies the persuadable fraction, where targeting concentrates responders above what random selection would achieve. A flat segment indicates patients whose outcome does not depend on treatment, either because they are controlled or uncontrolled regardless (sure things or lost causes), and a declining segment indicates a negative increment to the treated group relative to control, consistent with sleeping dogs.
	
	\subsection{Characterising the Persuadables}
	For the selected model on each cohort, we partition patients into predicted uplift groups and compare realised uplift (i.e. rate of BP control in PTEC-HT group minus rate of BP control in control group) with group-level confidence intervals. To further characterise features of the persuadable segment, we show the feature importance plots for each model and show the rates of BP control in PTEC-HT and control groups within bins of each top feature.
	
    From the top features identified by the uplift models, we also construct a handcrafted rule to estimate the individual uplift. Unlike the five learned models, this rule requires no model fitting and reflects criteria simple enough for direct use in clinical workflows. It is evaluated identically to the learned models, using its score to rank patients and computing AUQC on the same held-out test set.

	\begin{table}[htbp]
		\floatconts
		{tab:ate}%
		{\caption{Average treatment effect (ATE) of PTEC-HT on 6-month BP control in matched cohorts. Rows report the percentage of patients with controlled BP in each group at baseline and at 6 months. ATE measures the difference in 6-month control rates (PTEC-HT $-$ Control), expressed in percentage points (pp).}}%
		{		\setlength{\tabcolsep}{2pt}
			\begin{tabular}{@{}lcc@{}}
				\toprule
				& \bfseries Full & \bfseries Uncontrolled BP\\
				\midrule
				Matched pairs $n$ & 7{,}278 & 2{,}053\\
				PTEC-HT, baseline & 70.0\% & 0.0\%\\
				Control, baseline & 70.0\% & 0.0\%\\
				PTEC-HT, 6 months & 74.2\% & 62.6\%\\
				Control, 6 months & 69.0\% & 50.4\%\\
				ATE (pp) & +5.2  & +12.2 \\
				ATE 95\% CI (pp)  & 3.7 - 6.7 & 9.2 - 15.2 \\
				$p$-value & $3\times10^{-12}$ & $1\times10^{-15}$\\
				\bottomrule
		\end{tabular}}
	\end{table}

	\begin{figure*}[!t]
		\floatconts
		{fig:qini}
		{\caption{Qini curves for the five uplift models, evaluated on the held-out test set.}}
		{%
			\subfigure[Full cohort]{
				\includegraphics[width=0.3\linewidth]{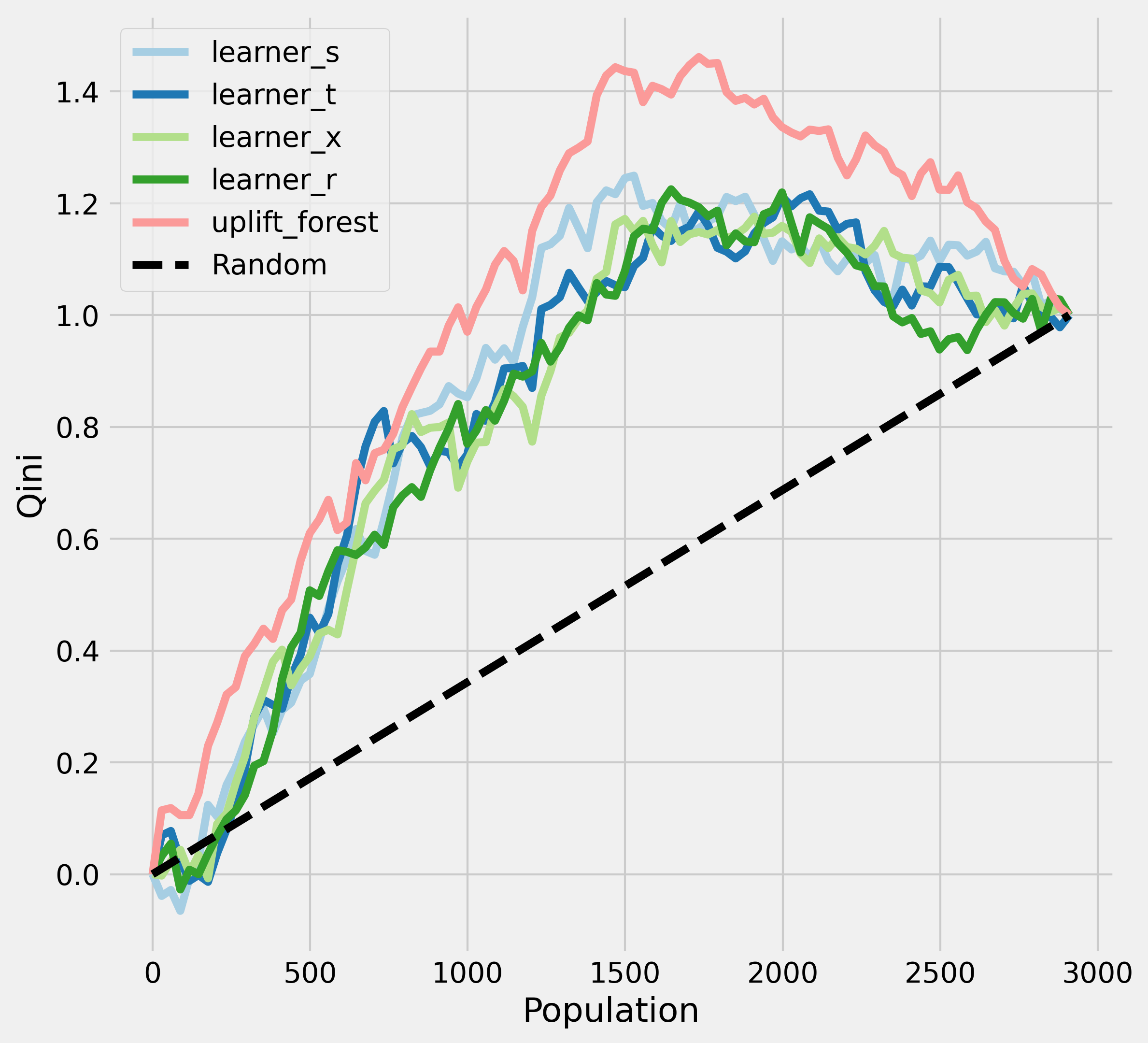}}%
			\qquad
			\subfigure[Uncontrolled BP cohort]{
				\includegraphics[width=0.3\linewidth]{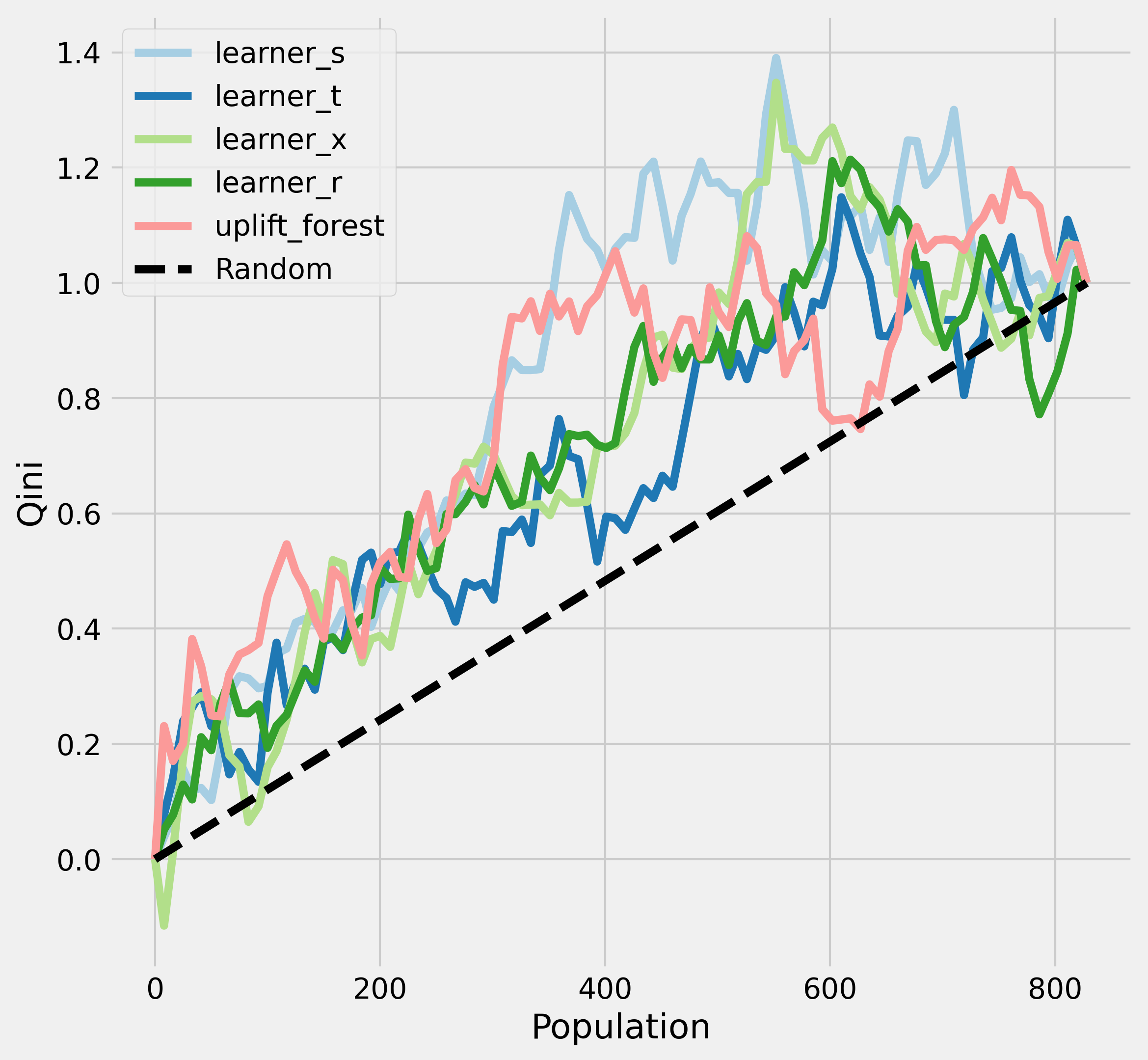}}
		}
	\end{figure*}
	
	\section{Results}
	\label{sec:results}
	
	\subsection{PTEC-HT Improves BP Control on Average}
	\label{sec:ate}
	
	\tableref{tab:ate} shows the results for average treatment effect. In the full matched cohort, both groups had equal BP control at baseline (70.0\%). After the 6-month intervention period, 74.2\% of PTEC-HT patients had controlled BP versus 69.0\% of matched controls, an absolute improvement of 5 percentage points (95\% CI 3.7 to 6.7 pp, $p = 3\times10^{-12}$). These estimates are directionally consistent with a separate real-world evaluation of the PTEC-HT scaling programme, which reported a 16 percentage point higher probability of BP control at 12 months, and 13.5 percentage points at 6 months \citep{wang2025economic}. The larger effect size in that study likely reflects differences in cohort definition, methodology and sample size.
	
	The effect is substantially larger where there is more room to improve. In the uncontrolled-baseline cohort, where all patients start uncontrolled, 62.6\% of PTEC-HT patients achieved control at 6 months versus 50.4\% of matched controls, a 12 percentage point improvement (95\% CI 9.2 to 15.2 pp, $p = 1\times10^{-15}$).

	\subsection{Model Selection by Qini Curves}
	\label{sec:qini}
	\tableref{tab:auqc} reports mean cross-validated AUQC across 5 folds of the training set for all five learners in both cohorts. On the full cohort, the uplift random forest achieved the highest mean AUQC. On the uncontrolled BP cohort, the S-learner achieved the highest mean AUQC. The same model also ranked highest among the learned models on the held-out test set in both cohorts. However, its $\pm 1$ SD band overlaps with those of all other learners, so the ranking is not decisive at these sample sizes and model comparisons are indicative only. Test AUQC exceeds CV AUQC for every model, most markedly in the uncontrolled BP cohort, whose test set has only about 410 matched pairs, so absolute values should not be compared between cross-validation and test.
	
	\figureref{fig:qini} shows the corresponding Qini curves on the held-out test set. All five learners lie clearly above the random-targeting diagonal, with tighter, earlier-separating curves in the full cohort and noisier, more overlapping curves in the smaller uncontrolled BP cohort, consistent with narrower AUQC margins in \tableref{tab:auqc}. The selected model in each cohort, the uplift forest and S-learner respectively, is also visually identifiable as the curve enclosing the largest area.
	
	\begin{table}[t]
		\floatconts
		{tab:auqc}%
		{\caption{Cross-validated (CV) AUQC (mean $\pm$ SD over the 5 folds of the training set) and held-out test AUQC. Bold marks the model selected by CV AUQC in each cohort.}}%
		{%
			\footnotesize
			\setlength{\tabcolsep}{4pt}
			\begin{tabular}{@{}lcccc@{}}
				\toprule
				& \multicolumn{2}{c}{\bfseries Full cohort} & \multicolumn{2}{c}{\bfseries Uncontrolled BP} \\
				\cmidrule(lr){2-3} \cmidrule(lr){4-5}
				\bfseries Model & \bfseries CV & \bfseries Test & \bfseries CV & \bfseries Test \\
				\midrule
				S-learner        & 0.30 $\pm$ 0.08 & 0.371 & \bfseries 0.11 $\pm$ 0.06 & 0.335 \\
				T-learner        & 0.32 $\pm$ 0.08 & 0.342 & 0.10 $\pm$ 0.05 & 0.171 \\
				X-learner        & 0.32 $\pm$ 0.06 & 0.333 & 0.09 $\pm$ 0.03 & 0.230 \\
				R-learner        & 0.31 $\pm$ 0.05 & 0.324 & 0.09 $\pm$ 0.08 & 0.214 \\
				Uplift forest    & \bfseries 0.41 $\pm$ 0.08 & 0.520 & 0.08 $\pm$ 0.04 & 0.281 \\
				\addlinespace
				Handcrafted rule & - & 0.455 & - & 0.393 \\
				\bottomrule
			\end{tabular}%
		}
	\end{table}

	\begin{figure*}[!t]
		\floatconts
		{fig:uplift_groups}
		{\caption{Realised uplift by predicted-uplift group (top 10\%, next 40\%, next 40\%, bottom 10\%) with group-level confidence intervals.}}
		{%
			\subfigure[Full cohort]{ 
				\includegraphics[width=0.35\linewidth]{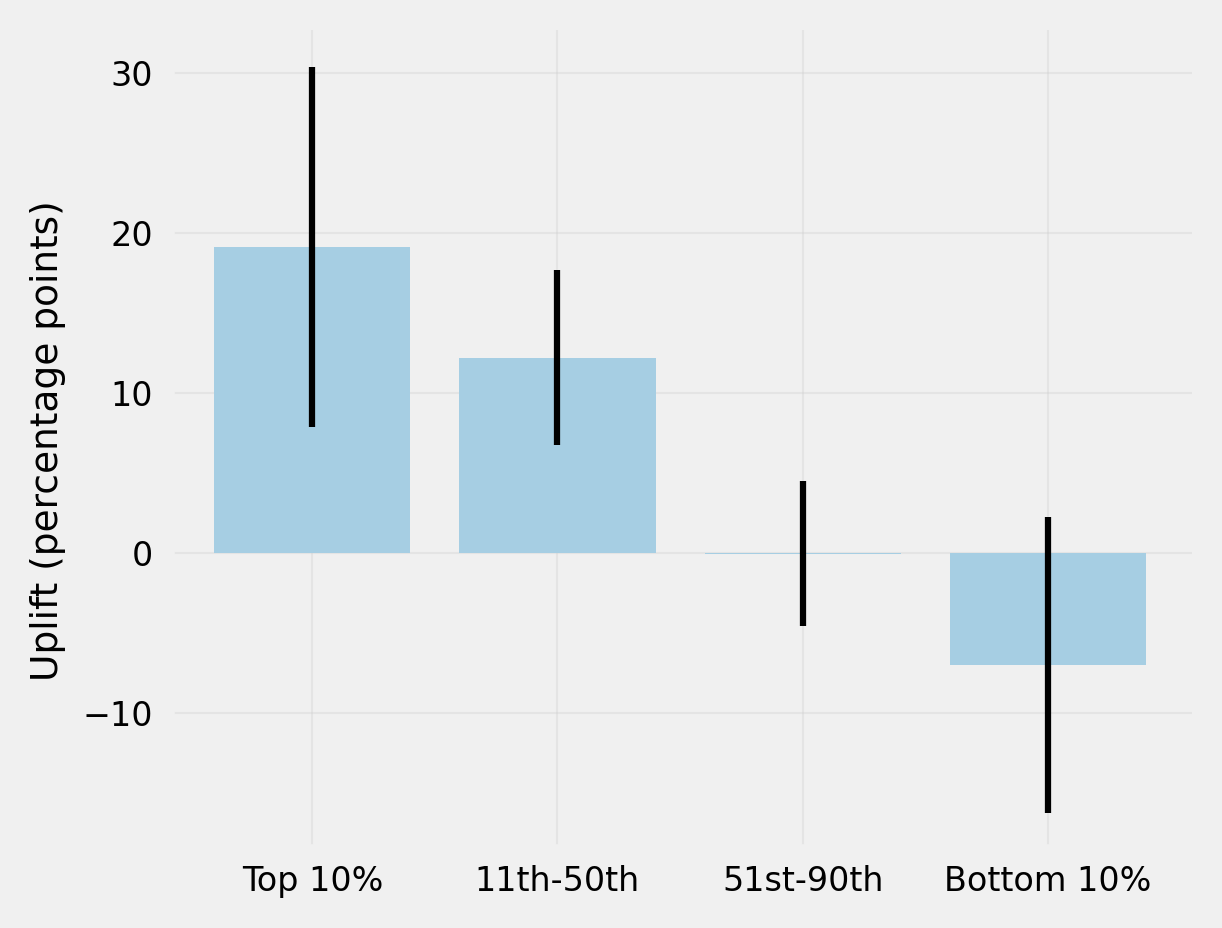}}%
			\qquad
			\subfigure[Uncontrolled BP cohort]{
				\includegraphics[width=0.35\linewidth]{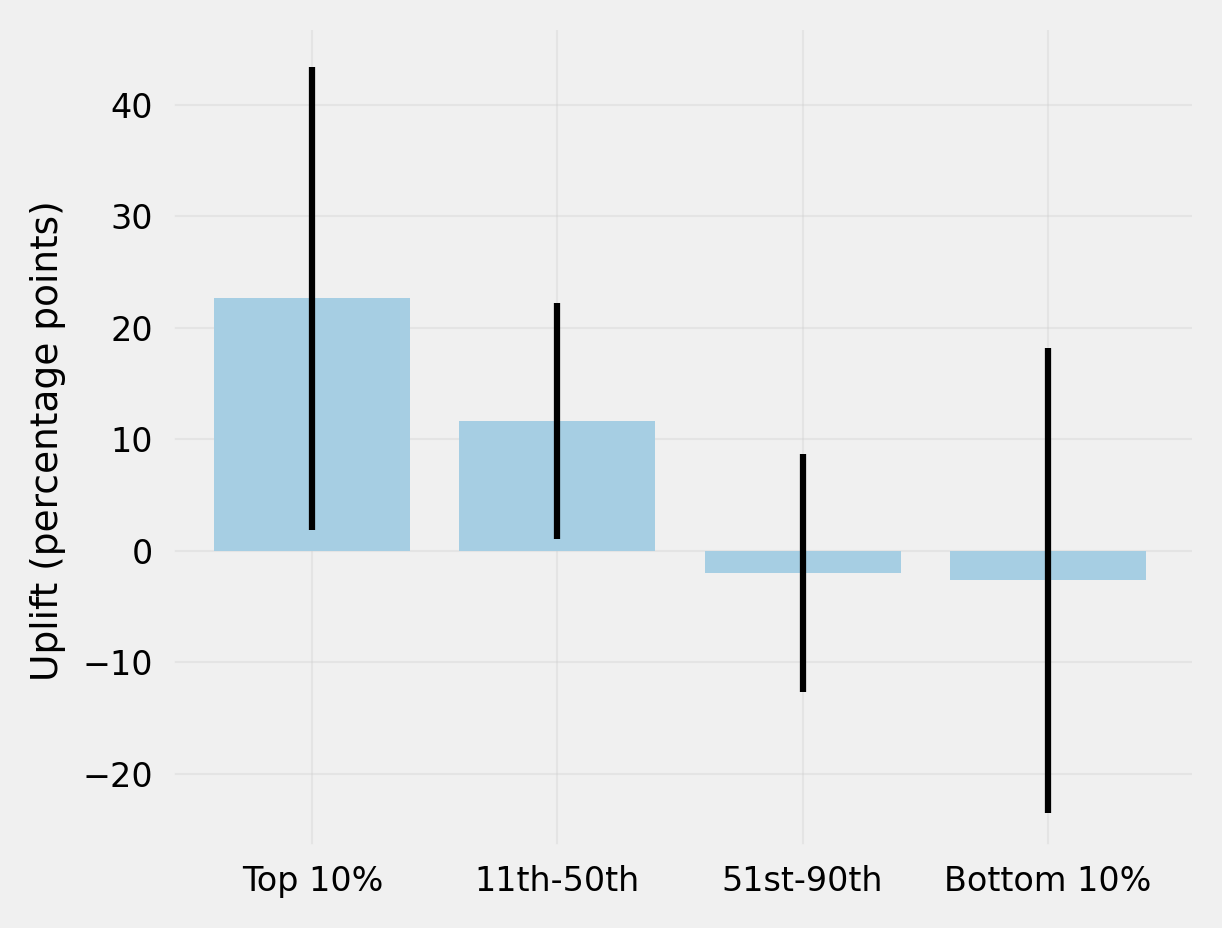}}
		}
	\end{figure*}
	
	\begin{figure*}[!t]
		\floatconts
		{fig:featimpt}
		{\caption{Feature importance plots for the uplift forest on full cohort and S-learner for uncontrolled BP cohort. The two models identified slightly different sets of top features.}}
		{%
			\subfigure[Full cohort]{
				\includegraphics[width=0.45\linewidth]{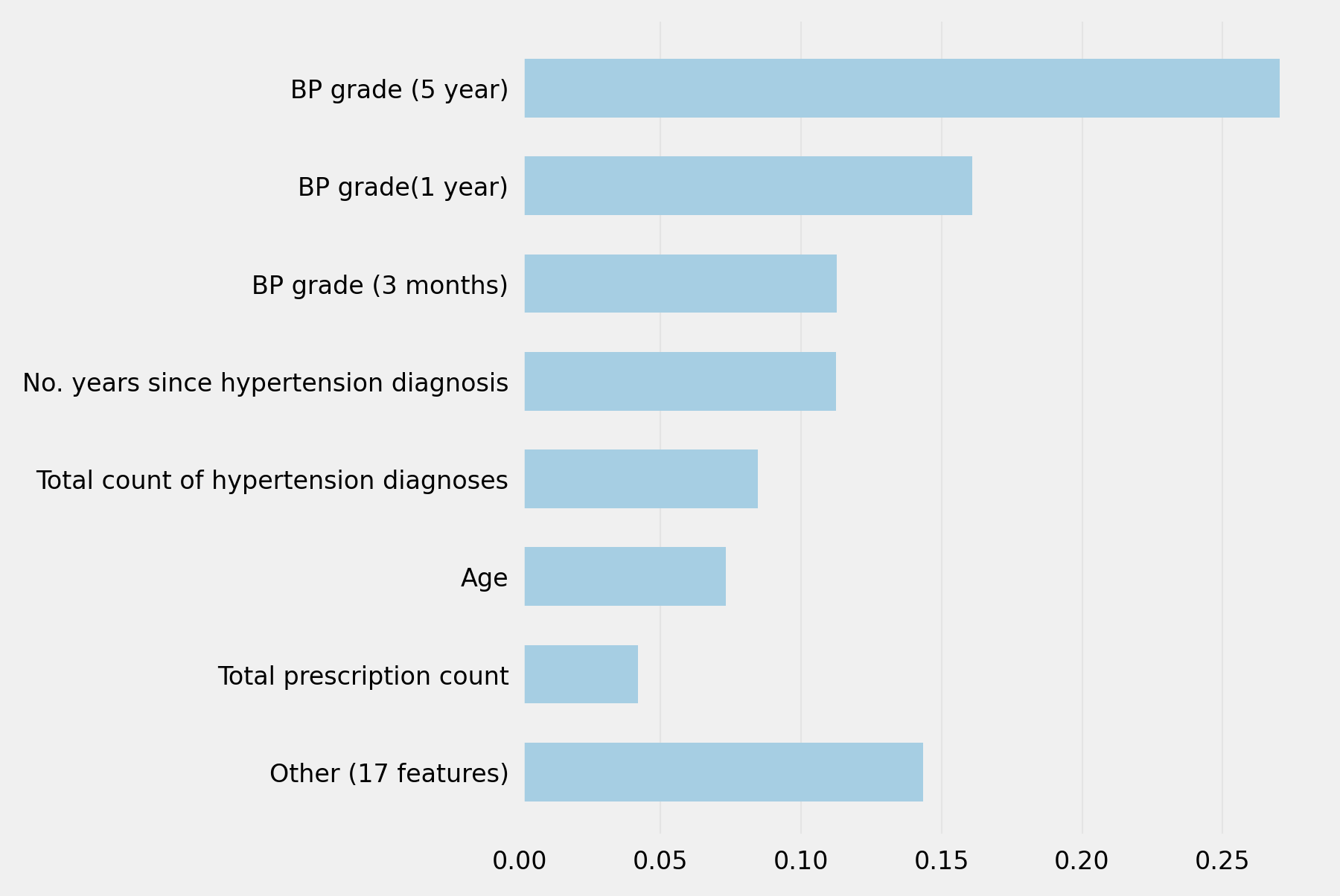}}%
			\qquad
			\subfigure[Uncontrolled BP cohort]{
				\includegraphics[width=0.45\linewidth]{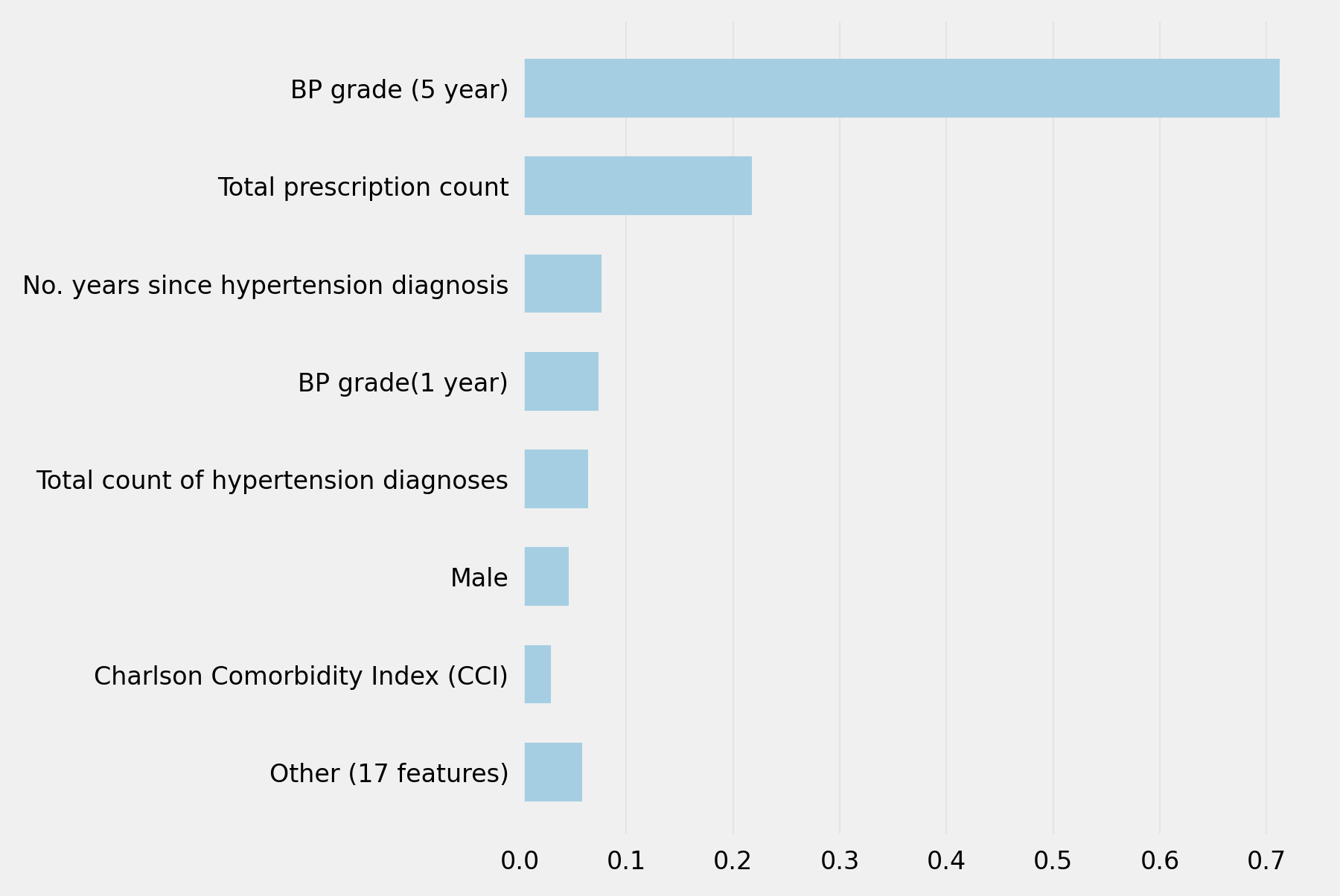}}
		}
	\end{figure*}

	\subsection{Realised Uplift by Predicted Uplift Group}
	\label{sec:deciles}
	
	We rank held-out test patients by predicted uplift and split them into four groups (top 10\%, next 40\%, next 40\% and bottom 10\%). In both cohorts realised uplift declines across the groups, and only the top two groups have confidence intervals that exclude zero (\figureref{fig:uplift_groups}). In the full cohort the top 10\% show 19 pp (95\% CI approximately 8--30 pp), with the lower bound above the overall ATE of 5 pp, followed by 12 pp in the next 40\%, near zero in the following 40\% and $-7$ pp in the bottom 10\%. The uncontrolled BP cohort shows the same ordering, with 23 pp in the top 10\% (95\% CI approximately 2--44 pp) against an overall ATE of 12 pp, 12 pp in the next 40\% and near-zero or slightly negative estimates below. Here the top-decile interval includes the overall effect, so we treat this concentration as suggestive. As a coarse calibration check, mean predicted uplift in each group follows the same ordering as realised uplift in both cohorts and lies within its 95\% CI, although predicted values span a narrower range (\appendixref{apd:calibration}).
	
	The bottom groups are less certain. The negative full-cohort bottom-decile estimate is consistent with the declining tail of the Qini curve (\figureref{fig:qini}), but its interval includes zero, so it is not conclusive evidence of a sleeping-dogs segment. In the uncontrolled BP cohort the bottom-group intervals are too wide to distinguish sleeping dogs from lost causes. 
	
	The group estimates are not confounding-adjusted. Within the top decile of the training set, several covariates used in the model show absolute standardised mean differences above 0.1, including antihypertensive medication count (SMD $-0.35$), hypertension diagnosis count (SMD $-0.34$) and years since diagnosis (SMD $-0.40$). In the held-out test set, where \figureref{fig:uplift_groups} is computed, all absolute SMDs are below 0.1 (\appendixref{apd:topdecile_smd}), although the top decile contains only about 290 patients so these values are noisy. We therefore cannot exclude that part of the observed 19 pp in the top 10\% of the full cohort reflects residual imbalance rather than genuine effect heterogeneity, and the direction of any resulting bias is unknown. Doubly-robust adjustment or subgroup-specific re-matching within the top decile would give confounding-adjusted estimates and is left to future work.

	\subsection{Who are the Persuadables?}
	\label{sec:persuadables}
	
	\figureref{fig:featimpt} shows feature importance for the uplift random forest (full cohort) and the S-learner (uncontrolled BP cohort). For the S-learner, feature importance is computed using the causalml library’s \texttt{get\_importance} function, which fits a LightGBM surrogate regressor on the predicted CATE and reports permutation importance on that surrogate. For the uplift forest, we report the built-in mean decrease in impurity (MDI) feature importance, which reflects which covariates most strongly partition patients by predicted treatment effect across the ensemble. In both cohorts, baseline BP grade measured over the 5-year window is the most important feature by a wide margin. The next-ranked features differ slightly between models: in the full cohort, the remaining BP grade windows (1-year, 3-month) and years since diagnosis follow closely behind, and in the uncontrolled BP cohort, total antihypertensive medication prescription count ranks second, well ahead of years since diagnosis and the 1-year BP grade window. Both models therefore converge on BP grade as the dominant driver of predicted uplift, with medication burden emerging as a secondary driver more prominently in the uncontrolled BP cohort.

	\figureref{fig:featgroup_4feat} shows the rate of BP control in the PTEC-HT and control groups within bins of the top features. The trends for the 3-month and 1-year baseline BP grades are similar to the 5-year grade and are shown in \appendixref{apd:featgroup_all}. In both cohorts baseline BP grade gives the clearest pattern. The gap between PTEC-HT and control patients is near zero for patients with normal or high-normal BP and widens with severity, reaching roughly 20 percentage points at Grade 2. At Grade 3 sample sizes are small and the confidence intervals overlap.
	
	Years since diagnosis and total diagnosis count show a related but distinct pattern, again in both cohorts. Patients with no prior hypertension diagnosis, newly diagnosed at or near PTEC-HT enrolment, show a substantially larger gap than patients with any diagnosis history. Among established patients the gap is smaller and fluctuates, and it narrows further among the most established. This is a step change between newly diagnosed patients and everyone else and not a smooth decline with disease duration. Medication count is less informative. In the full cohort it shows a weaker, noisier relationship with no clear monotonic trend, whereas in the uncontrolled BP cohort the gap is largest among patients with 1--10 prescriptions and smallest among those with 21 or more, consistent with medication count ranking second for the S-learner in this cohort.

		\begin{figure}[htbp]
		\floatconts
		{fig:featgroup_4feat}
		{\caption{Rate of BP control in the PTEC-HT and control groups within bins of each top feature for the full and uncontrolled BP cohort. The BP grades are N: normal BP, HN: high-normal BP, G1-G3: grades 1, 2, 3 hypertension.}}
		{%
			\subfigure[Full cohort]{
					\label{fig:featgroup_full_4feat}
				\includegraphics[width=0.95\linewidth]{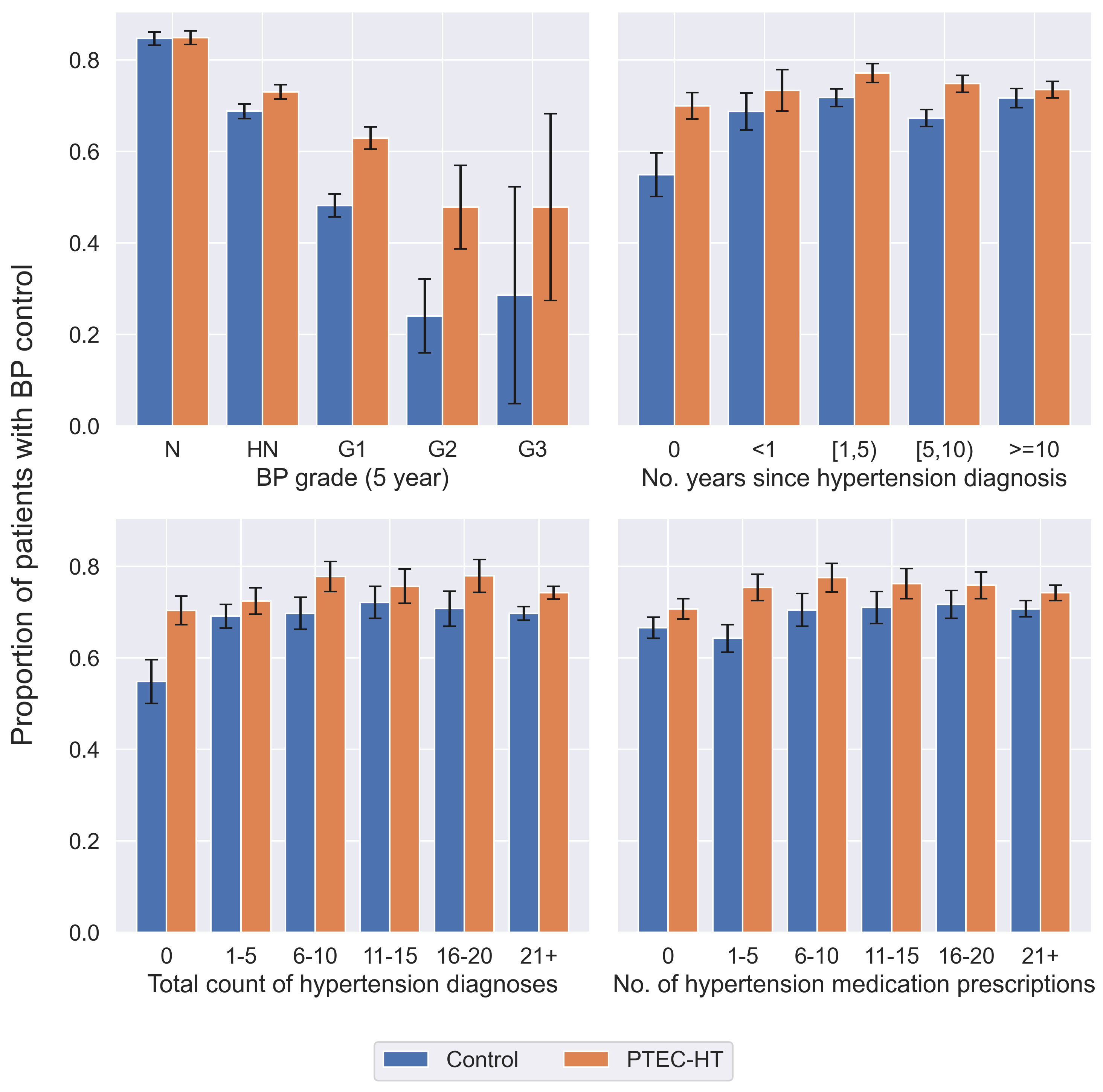}}%
			\par\medskip
			\subfigure[Uncontrolled BP cohort]{
					\label{fig:featgroup_uncontrolled_4feat}
				\includegraphics[width=0.95\linewidth]{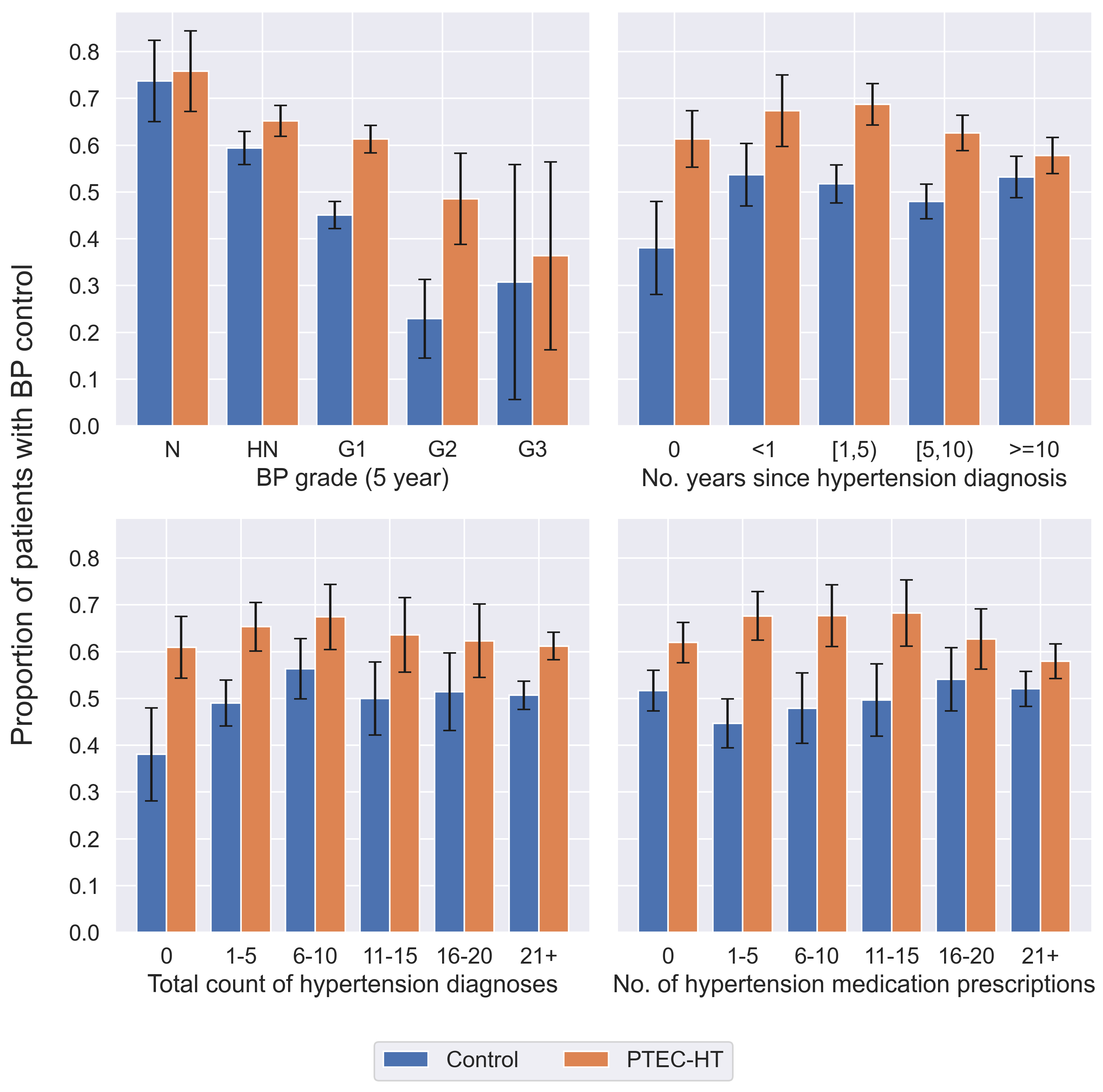}}
		}
	\end{figure}

	Years since diagnosis, number of hypertension diagnoses and number of antihypertensive prescriptions are mechanically correlated because a longer disease history provides more opportunities for events to be recorded (\appendixref{apd:featcorr}). We therefore read them as a composite measure of disease history and management burden and not as separate effects. \figureref{fig:featgroup_4feat} shows the same pattern without relying on model-fitted importance.
	
	We hypothesise that the programme is most beneficial for patients with poorly controlled but recently diagnosed hypertension, whose disease is not yet heavily managed and who plausibly have the most to gain from structured self-monitoring and timely titration. In uplift terms it is least beneficial for long-standing, already-controlled patients or long-standing uncontrolled patients on established, high-burden regimens. This profile echoes an individual-patient-data meta-analysis of 25 randomised trials of BP self-monitoring, in which self-monitoring was most effective in patients with fewer antihypertensive medications and higher baseline systolic BP \citep{tucker2017}. That analysis used a meta-regression across predominantly Western trials of self-monitoring whereas ours applies an uplift model to a binary control outcome in a single national telemonitoring programme, so this is convergent evidence and not a replication, but it is a useful external check on the clinical plausibility of the learned profile.

	\paragraph{Comparison with Handcrafted Rule} 
	Since the learned uplift models require model inference to score new patients, we also construct a handcrafted rule from their top features that care teams could apply directly during enrolment without running a model. The handcrafted rule thresholds were determined from training-set bin plots only to avoid data leakage to the test set. For the full cohort, the score is the sum of baseline BP grade (5-year window, scored 0–4 from Normal to Grade 3) and an indicator for no prior hypertension diagnosis (newly diagnosed at or near enrolment). For the uncontrolled BP cohort, the score additionally includes an indicator for 1–10 antihypertensive prescriptions. 
	
	On the independent test set, the handcrafted rule outperformed all five learned models on the uncontrolled BP cohort (test AUQC 0.393 vs. 0.171--0.335) but underperformed the uplift random forest on the full cohort (0.455 vs. 0.520; \tableref{tab:auqc}). This split likely reflects sample size. The uncontrolled BP cohort has under a third of the matched pairs of the full cohort and CV AUQC values roughly a quarter as large, so there is less signal for five separately-tuned learners to exploit without overfitting whereas a 2--3 feature rule generalises more reliably. Because the rule is built from the uplift models' own top features, its performance corroborates their feature importance rankings and does not undermine the value of uplift modelling.

	\section{Discussion}
	\label{sec:discussion}

	\paragraph{Operational Implications} Our analysis suggests that uplift scores could be used to rank candidates and concentrate onboarding capacity on the top segment through targeted outreach. However, this analysis differs from predicting which patients will accept an offer to enrol, which requires acquisition-funnel data. Whether a fitted model is worth deploying over simple criteria depends on sample size (\sectionref{sec:persuadables}). We see the models' main contribution as a discovery tool, identifying the features that drive treatment-effect heterogeneity so that this structure can be distilled into a transparent rule for operational use where interpretability and simplicity matter. A fitted model is likely to add the most direct value at larger enrolment scales whereas simple criteria may generalise at least as well once the structure is known and samples are small.

	\paragraph{Clinical Need and Predicted Benefit} 
	Predicted benefit and clinical need can point in different directions. Patients with long-standing uncontrolled hypertension or a high medication burden have lower estimated uplift from telemonitoring (\sectionref{sec:persuadables}), which does not mean they need less care. For these patients telemonitoring alone may be insufficient and complementary interventions such as dietary counselling, weight management or supervised exercise could be considered, although we have not evaluated these here. Uplift scores should therefore be used to prioritise among clinically eligible candidates and not to replace clinical judgement. A multi-arm extension of the framework, comparing several intervention types at once, is a natural direction for future work.

	\paragraph{Limitations}
	 First, this is an observational study and propensity score matching balances observed covariates but cannot rule out unmeasured confounding (e.g. health literacy, motivation, digital access). We computed E-values for the average treatment effects \citep{vanderweele2017sensitivity}. For the full cohort the risk ratio (RR) of BP control is 1.08 and the E-value is 1.36; for the uncontrolled BP cohort, RR is 1.24 and E-value is 1.79, meaning an unmeasured confounder would need to be associated with both enrolment and BP control by at least these risk ratios, beyond the matched covariates, to fully explain the observed effects. These values are modest, so confounding of plausible strength could account for the full-cohort effect, while the uncontrolled-cohort effect is somewhat more robust. They apply to the average effects only and we have not performed a sensitivity analysis specific to the uplift rankings.
	
	Second, the comparison groups are not indexed at equivalent points in care. Treated patients are indexed at PTEC-HT registration, which is likely to coincide with a clinic visit and possible medication adjustment, whereas control patients are assigned a random index date drawn from the range of registration dates. This time-zero mismatch could inflate the estimated effect, particularly for patients with high BP or recent diagnosis. In the full cohort, medication recency (median days since last antihypertensive prescription, treated: 526 vs control: 532) and diagnosis recency (treated: 163 vs control: 172 days) were similar between matched arms, but the most recent clinic BP preceded the index date by a median of 38 days in PTEC-HT patients and 85 days in controls, consistent with registration coinciding with a visit. We therefore cannot exclude residual bias from differing clinical contact at the index date. An active-comparator design (patients offered but not enrolled in PTEC-HT) or controls anchored to a clinical event would address this in prospective evaluation.
	
	Third, the outcome requires a clinic BP within 6 months, so both cohorts condition on healthcare contact during follow-up. Only 35\% of sampled control candidates met this criterion compared with 51\% of PTEC-HT patients, which is expected as programme reminders and teleconsultations promote clinic contact. Both arms are filtered before analysis, but because participation itself changes the probability of being observed, this filtering may bias the estimated effects in an unknown direction and not only limit generalisability. Included patients in both arms were older with more diagnoses, prescriptions and comorbidity than excluded patients, so the analysis population skews towards more established and more actively managed patients and our findings may not extend to those excluded. Likewise, our estimates describe matched patients and not all enrolees. Patients unmatched within caliper resemble the higher-uplift profile, so the patients for whom the programme appears most promising may be under-represented, and their treatment effect cannot be estimated without comparable controls.
	
	Lastly, white coat hypertension, where clinic BP is usually higher than BP measured at home, has not been accounted for. Clinic BP is the outcome in both arms but frequent home monitoring may change how patients respond to clinic measurement, and we lack concurrent home and clinic readings for the control arm to quantify this.

	\section{Conclusion and Future Work}
	\label{sec:conclusion}
	
	The PTEC-HT home BP telemonitoring programme improves 6-month BP control on average. Uplift modelling suggests that the benefit is heterogeneous, with larger realised uplift among patients with higher baseline BP grade, more recent diagnosis and lower medication burden, although these estimates are imprecise and not fully confounding-adjusted. A simple handcrafted rule built from these features outperformed all learned models on the uncontrolled BP cohort, so such baseline characteristics may be usable directly to inform targeted recruitment and to direct limited care team capacity toward patients likely to benefit.
	
	Future work includes five directions. First, incorporating acquisition-funnel data (offer dates, patient response) would allow enrolment propensity to be modelled jointly with treatment benefit. Second, prospective validation via a staged or randomised rollout offering PTEC-HT preferentially to high-uplift patients would test whether this retrospective analysis holds under real-world constraints. Third, comparing the uplift models against a logistic-regression risk-stratification score that predicts BP control from baseline covariates alone would show whether targeting gains come from genuine treatment-effect heterogeneity or merely from ranking patients by how controllable their BP is regardless of the programme. Fourth, a multi-arm extension comparing telemonitoring with complementary interventions such as dietary counselling, weight management and exercise would allow patients to be matched to the intervention most likely to benefit them. Lastly, PTEC covers chronic conditions beyond hypertension, and extending the framework to its other modules would show whether the persuadable profile identified here reflects a general pattern in remote chronic disease management.
	
	
	\bibliography{ref}

\appendix
	
\section{Cohort Construction and Matching}

\subsection{Cohort Flow Diagram}\label{apd:flow}
\figureref{fig:flow} traces how each cohort was derived. Of 16,694 registered patients, 8,563 (51\%) had a clinic BP in the 6-month follow-up window, yielding 7,278 matched pairs in the full cohort and 2,053 matched pairs from 2,976 eligible patients in the uncontrolled BP cohort.

\subsection{Covariate Balance After Matching}\label{apd:all_smd}
\tableref{tab:smd_matched} shows the standardised mean differences of covariates after matching in the full and uncontrolled BP cohorts. All covariates have $|\mathrm{SMD}| < 0.1$. 
		
\subsection{Unmatched vs. Matched Patients}\label{apd:matched_vs_unmatched}
\tableref{tab:matched-unmatched} reports standardised mean differences between unmatched and matched PTEC-HT patients in the full and uncontrolled BP cohorts, computed as unmatched minus matched. The differences in age, years since diagnosis and baseline BP grade are summarised in \sectionref{sec:psm}. Unmatched patients were also less often in the Merdeka and Pioneer Generation CHAS tiers (MG, MGA, PG and PGA, SMD -0.20 to -0.31) and more often in tier C or without a CHAS tier (0.19 to 0.32). The Merdeka and Pioneer Generation schemes are restricted to older birth cohorts, so this is consistent with the age difference. Household size and Charlson Comorbidity Index differed less.

\section{Features and Implementation}
\subsection{Feature Definitions}\label{apd:model_featdescription}	
\tableref{tab:feature-defs} provides full definitions and derivation logic for the baseline covariates used in propensity score matching (\sectionref{sec:psm}) and uplift modelling (\sectionref{sec:methods_upliftmodels}). Baseline BP grade was encoded as one-hot categorical variables across three lookback windows (3 months, 1 year, 5 years) for propensity score estimation and as an ordinal scalar (0--4, Normal to Grade 3) for the uplift models, reflecting its inherent severity ordering. Antihypertensive medication prescriptions were identified using a lookup of ATC codes spanning diuretics, beta blockers, calcium channel blockers and RAAS agents (including ACE inhibitors and angiotensin receptor blockers), covering both single-agent and multi-agent combination products.

\subsection{Pseudocode for Matching and Uplift Modelling}\label{apd:pseudocode}	
\algorithmref{alg:matching} summarises cohort construction and propensity score matching (\sectionref{sec:psm}) and \algorithmref{alg:uplift} summarises uplift model training, selection and evaluation (\sectionref{sec:methods,sec:results}). Both are run separately for the full and uncontrolled BP cohorts. Train-test splits and cross-validation folds are formed at the matched-pair level whereas ranking and evaluation are at the patient level. Uplift learners are implemented with \texttt{causalml} \citep{chen2020causalml} and the propensity model and nearest-neighbour search with \texttt{scikit-learn} \citep{scikit-learn}.

\subsection{Hyperparameters for Uplift Models}\label{apd:model_hyperparam}
All uplift models were implemented using the \texttt{causalml} Python library, which provides the S-, T-, X-, and R-learner meta-learner classes (\texttt{BaseSClassifier}, \texttt{BaseTClassifier}, \texttt{BaseXClassifier}, \texttt{BaseRClassifier}) and the \texttt{UpliftRandomForestClassifier} used in this work. Base learners and hyperparameters were selected separately for the full and uncontrolled BP cohorts via cross-validation, spanning logistic regression, random forest and L2-regularised linear/SGD classifiers. Full configurations are reported in \tableref{tab:hyperparams}.

\section{Supplementary Results}
\subsection{Calibration of Predicted Uplift}
\label{apd:calibration}
\figureref{fig:calibration} compares mean predicted uplift with realised uplift in the four predicted-uplift groups of Figure~\ref{fig:uplift_groups} on the held-out test set, for the selected model in each cohort. In both cohorts mean predicted uplift decreases across the groups in the same order as realised uplift and lies within the 95\% confidence interval of realised uplift in every group. Predicted uplift is more compressed than realised uplift. In the full cohort it ranges from $-1$ pp in the bottom group to 18 pp in the top group whereas realised uplift ranges from $-7$ pp to 19 pp. In the uncontrolled BP cohort predicted uplift ranges from 3 pp to 10 pp whereas realised uplift ranges from about $-3$ pp to 23 pp, so the model does not reproduce the near-zero or negative uplift observed in the lower groups.

The confidence intervals are wide, particularly in the smaller uncontrolled BP cohort, so agreement within the intervals is a coarse check on ordering and rough scale and not a precise calibration assessment. The intervals also describe sampling variability in realised uplift only and not uncertainty in the predicted scores, which we leave to future work.

\subsection{Covariate Balance for Top-Decile Subgroup}\label{apd:topdecile_smd}
\tableref{tab:topdecile_smd} shows the standardised mean differences of covariates for the top-decile predicted uplift subgroup. Several covariates used in the model show standardised mean differences of above 0.1 in the training set, including antihypertensive medication count (SMD -0.35), hypertension diagnosis count (SMD -0.34) and years since diagnosis (SMD -0.40).

\subsection{Correlation Among Key Features}\label{apd:featcorr}
Years since diagnosis, number of hypertension diagnoses and number of antihypertensive prescriptions are mechanically correlated, since a longer disease history provides more opportunities for subsequent diagnoses and prescriptions to be recorded. \tableref{tab:corr} shows the pairwise correlations among these three features. Spearman correlation was used as number of hypertension diagnoses and number of antihypertensive prescriptions are right-skewed counts.

\subsection{Rate of BP Control by Feature Bins}\label{apd:featgroup_all}
\figureref{fig:featgroup} reports the rate of BP control in the PTEC-HT and control groups within bins of baseline BP grade measured over the 3-month and 1-year windows, complementing the 5-year window shown in the main text. Both cohorts show the same widening pattern with increasing BP grade severity described in 	\sectionref{sec:persuadables}, with the gap near zero for normal or high-normal BP and increasing at higher grades, subject to wider confidence intervals at Grade 3 due to small subgroup sizes. These trends are consistent with the 5-year window patterns discussed in \sectionref{sec:persuadables} and are reported here for completeness.

	\begin{figure}[t]
		\centering
		\begin{tikzpicture}[
			font=\footnotesize, node distance=5mm,
			box/.style={draw, align=center, text width=0.42\linewidth, inner sep=3pt},
			wide/.style={box, text width=0.80\linewidth},
			excl/.style={box, dashed},
			wexcl/.style={wide, dashed},
			lab/.style={font=\footnotesize\bfseries},
			arr/.style={-{Stealth[length=4pt]}}
			]
			\node[wide] (reg) {PTEC-HT patients registered\\August 2020 to May 2024 ($n = 16{,}694$)};
			\node[wexcl, below=of reg] (nobp) {Excluded: no clinic BP in the 6 months after enrolment ($n = 8{,}131$)};
			\node[wide, below=of nobp] (elig) {Eligible patients with a clinic BP in the 6 months after enrolment ($n = 8{,}563$)};
			
			\node[lab, anchor=north] (fl) at ([shift={(-0.24\linewidth,-5mm)}]elig.south) {Full cohort};
			\node[lab, anchor=north] (ul) at ([shift={(0.24\linewidth,-5mm)}]elig.south) {Uncontrolled BP cohort};
			
			\node[excl, below=of ul] (nu) {Excluded: not uncontrolled at 1-year baseline ($n = 5{,}587$)};
			\node[box, below=of nu] (u2) {Uncontrolled at baseline ($n = 2{,}976$)};
			\node[excl, below=of u2] (unm2) {Excluded: unmatched within caliper ($n = 923$)};
			\node[box, below=of unm2] (m2) {Matched pairs ($n = 2{,}053$)};
			
			\node[excl] (unm1) at (fl |- unm2) {Excluded: unmatched within caliper ($n = 1{,}285$)};
			\node[box] (m1) at (fl |- m2) {Matched pairs ($n = 7{,}278$)};
			
			\draw[arr] (reg) -- (nobp);
			\draw[arr] (nobp) -- (elig);
			\draw[arr] (elig.south) -- ++(0,-2.5mm) -| (fl.north);
			\draw[arr] (elig.south) -- ++(0,-2.5mm) -| (ul.north);
			\draw[arr] (fl) -- (unm1);
			\draw[arr] (unm1) -- (m1);
			\draw[arr] (ul) -- (nu);
			\draw[arr] (nu) -- (u2);
			\draw[arr] (u2) -- (unm2);
			\draw[arr] (unm2) -- (m2);
		\end{tikzpicture}
		\caption{Cohort flow diagram. Dashed boxes denote exclusions. The uncontrolled BP cohort is drawn from the same eligible patients as the full cohort and matching was performed separately within each cohort. Matching was 1:1 so each cohort contains twice as many patients as matched pairs.}
		\label{fig:flow}
	\end{figure}

\begin{table}[t]
	\centering
	\footnotesize
	\caption{Covariate balance after matching (standardised mean differences) in the full and uncontrolled BP cohorts. All covariates have $|\mathrm{SMD}| < 0.1$. A dash indicates the category is empty in that cohort.}
	\label{tab:smd_matched}
	\begin{tabular}{lrr}
		\toprule
		& \multicolumn{2}{c}{\textbf{SMD}} \\
		\cmidrule(l){2-3}
		\textbf{Covariate} & \textbf{Full} & \textbf{Uncontrolled BP} \\
		\midrule
		Age                                  & 0.007    & 0.012    \\
		Male                                 & $-0.017$ & $-0.024$ \\
		Years since diagnosis                & 0.000    & 0.011    \\
		Total diagnoses         & $-0.004$ & 0.007    \\
		Total prescriptions & $-0.009$ & 0.016    \\
		Household size                       & $-0.009$ & $-0.019$ \\
		Total CCI score                      & 0.012    & $-0.009$ \\
		\midrule
		\multicolumn{3}{l}{\textit{BP grade (3-month)}} \\
		\quad Normal                         & 0.008    & $-0.011$ \\
		\quad High-normal                    & 0.000    & $-0.010$ \\
		\quad Grade 1                        & $-0.010$ & 0.026    \\
		\quad Grade 2                        & 0.000    & 0.000    \\
		\quad Grade 3                        & 0.004    & $-0.042$ \\
		\addlinespace
		\multicolumn{3}{l}{\textit{BP grade (1-year)}} \\
		\quad Normal                         & 0.012    & --       \\
		\quad High-normal                    & $-0.008$ & --       \\
		\quad Grade 1                        & $-0.007$ & 0.024    \\
		\quad Grade 2                        & 0.008    & $-0.009$ \\
		\quad Grade 3                        & $-0.005$ & $-0.036$ \\
		\addlinespace
		\multicolumn{3}{l}{\textit{BP grade (5-year)}} \\
		\quad Normal                         & 0.008    & $-0.005$ \\
		\quad High-normal                    & $-0.002$ & $-0.016$ \\
		\quad Grade 1                        & $-0.008$ & 0.019    \\
		\quad Grade 2                        & 0.003    & 0.000    \\
		\quad Grade 3                        & $-0.002$ & $-0.010$ \\
		\midrule
		\multicolumn{3}{l}{\textit{CHAS subsidy tier}} \\
		\quad A                              & 0.003    & $-0.028$ \\
		\quad B                              & 0.007    & $-0.019$ \\
		\quad C                              & $-0.017$ & 0.009    \\
		\quad MG                             & 0.005    & 0.001    \\
		\quad MGA                            & 0.000    & 0.016    \\
		\quad MGB                            & 0.002    & $-0.002$ \\
		\quad MGC                            & $-0.006$ & 0.013    \\
		\quad PG                             & $-0.001$ & 0.004    \\
		\quad PGA                            & 0.013    & $-0.013$ \\
		\quad PGB                            & 0.004    & 0.016    \\
		\quad PGC                            & $-0.004$ & 0.005    \\
		\quad None                           & $-0.003$ & 0.011    \\
		\bottomrule
	\end{tabular}
\end{table}


\begin{table}[t]
	\centering
	\footnotesize
	\caption{Standardised mean differences between unmatched and matched PTEC-HT patients (unmatched minus matched) in the full and uncontrolled BP cohorts. A dash indicates the category is empty in that cohort.}
	\label{tab:matched-unmatched}
	\begin{tabular}{lrr}
		\toprule
		& \multicolumn{2}{c}{\textbf{SMD}} \\
		\cmidrule(l){2-3}
		\textbf{Covariate} & \textbf{Full} & \textbf{Uncontrolled BP} \\
		\midrule
		Age                                  & $-0.597$ & $-0.677$ \\
		Male                                 & 0.333    & 0.243    \\
		Years since diagnosis                & $-0.417$ & $-0.597$ \\
		Total diagnoses                      & $-0.141$ & $-0.344$ \\
		Total prescriptions                  & $-0.080$ & $-0.298$ \\
		Household size                       & 0.163    & 0.189    \\
		Total CCI score                      & $-0.143$ & $-0.113$ \\
		\midrule
		\multicolumn{3}{l}{\textit{BP grade (3-month)}} \\
		\quad Normal                         & $-0.579$ & $-0.103$ \\
		\quad High-normal                    & $-0.225$ & $-0.254$ \\
		\quad Grade 1                        & 0.393    & $-0.155$ \\
		\quad Grade 2                        & 0.399    & 0.234    \\
		\quad Grade 3                        & 0.180    & 0.194    \\
		\addlinespace
		\multicolumn{3}{l}{\textit{BP grade (1-year)}} \\
		\quad Normal                         & $-0.583$ & --       \\
		\quad High-normal                    & $-0.192$ & --       \\
		\quad Grade 1                        & 0.456    & $-0.301$ \\
		\quad Grade 2                        & 0.339    & 0.250    \\
		\quad Grade 3                        & 0.147    & 0.154    \\
		\addlinespace
		\multicolumn{3}{l}{\textit{BP grade (5-year)}} \\
		\quad Normal                         & $-0.504$ & $-0.237$ \\
		\quad High-normal                    & $-0.163$ & $-0.266$ \\
		\quad Grade 1                        & 0.474    & $-0.135$ \\
		\quad Grade 2                        & 0.408    & 0.323    \\
		\quad Grade 3                        & 0.195    & 0.241    \\
		\midrule
		\multicolumn{3}{l}{\textit{CHAS subsidy tier}} \\
		\quad A                              & 0.045    & 0.089    \\
		\quad B                              & 0.134    & 0.198    \\
		\quad C                              & 0.302    & 0.194    \\
		\quad MG                             & $-0.212$ & $-0.202$ \\
		\quad MGA                            & $-0.309$ & $-0.282$ \\
		\quad MGB                            & $-0.136$ & $-0.139$ \\
		\quad MGC                            & $-0.167$ & $-0.175$ \\
		\quad PG                             & $-0.222$ & $-0.226$ \\
		\quad PGA                            & $-0.244$ & $-0.294$ \\
		\quad PGB                            & $-0.097$ & $-0.078$ \\
		\quad PGC                            & $-0.080$ & $-0.099$ \\
		\quad None                           & 0.236    & 0.322    \\
		\bottomrule
	\end{tabular}
\end{table}

	\begin{table*}[htbp]
		\centering
		\caption{Definitions and derivation of baseline covariates used for propensity score matching and uplift modelling.}
		\label{tab:feature-defs}
		{%
			\begin{tabular}{@{}p{0.27\linewidth}p{0.69\linewidth}@{}}
				\toprule
				\textbf{Feature} & \textbf{Definition} \\
				\midrule
				Age & Age at PTEC-HT registration date (treated) or assigned observation date (control) \\
				\addlinespace
				Gender & Male vs.\ female \\
				\addlinespace
				Years since diagnosis & No.\ of years from first coded hypertension diagnosis to observation date \\
				\addlinespace
				No.\ hypertension diagnoses & Total number of coded hypertension diagnosis events prior to observation date \\
				\addlinespace
				No.\ antihypertensive prescriptions & Total number of antihypertensive medication prescriptions prior to observation date, defined via ATC codes for diuretics (C03), beta blockers (C07), calcium channel blockers (C08) and RAAS agents/ARBs (C09) including single- and multi-agent combination products \\
				\addlinespace
				BP grade (3 months) & Mean clinic BP over the 3 months preceding observation, graded: Normal, High-normal, Grade 1,  Grade 2, Grade 3 \\
				\addlinespace
				BP grade (1 year) & As above, 1-year lookback window \\
				\addlinespace
				BP grade (5 year) & As above, 5-year lookback window \\
				\addlinespace
				Household size & Number of individuals registered at the same household address \\
				\addlinespace
				Charlson Comorbidity Index (CCI) & Weighted comorbidity index computed from coded diagnoses \\
				\addlinespace
				CHAS subsidy tier & Community Health Assist Scheme subsidy tier at observation date: A, B, C, MG, MGA, MGB, MGC, PG, PGA, PGB, PGC, None \\
				\bottomrule
			\end{tabular}%
		}
	\end{table*}

	\begin{algorithm*}[p]
		\caption{Cohort construction and propensity score matching \\ (run separately for the full and uncontrolled BP cohorts)}
		\label{alg:matching}
		\DontPrintSemicolon
		\LinesNotNumbered
		\KwIn{PTEC-HT registrants $\mathcal{T}$; non-PTEC-HT patients $\mathcal{C}_0$ aged 20 to 80 with a coded hypertension diagnosis, excluding pregnancy, atrial fibrillation, kidney failure and heart failure}
		$\mathcal{C} \gets$ random sample of 50,000 patients from $\mathcal{C}_0$\; \\
		\textbf{Index date:} registration date for $i \in \mathcal{T}$; a random date between January 2021 and May 2024 for $i \in \mathcal{C}$\; \\
		\textbf{Covariates:} compute all covariates $X_i$ from records before the index date\; \\
		\textbf{Outcome:} $Y_i \gets 1$ if mean clinic BP in the 180 days after the index date is below 140/90 mmHg and 0 otherwise; drop patients with no clinic BP in this window\; \\
		\textbf{Missing data:} set missing baseline clinic BP measurement to Normal grade BP; set missing counts (number of hypertension diagnoses, number of antihypertensive medications) and years since diagnosis to zero\; \\
		\If{uncontrolled BP cohort}{
			keep patients with uncontrolled 1-year baseline BP grade 
		}
		Fit logistic regression of treatment on standardised numeric and one-hot categorical covariates to obtain $\hat{e}(x)$ \\
		$r \gets 0.01 \times \mathrm{SD}(\hat{e}(x))$ among controls \\
		\For{each treated patient $i$ in turn}{
			$\mathcal{N}_i \gets$ unmatched controls $j$ with $|\hat{e}(x_j) - \hat{e}(x_i)| \le r$, sorted by distance \\
			\eIf{$\mathcal{N}_i \neq \emptyset$}{
				match $i$ to the first element of $\mathcal{N}_i$ (no replacement) 
			}{
				leave $i$ unmatched 
			}
		}
		\textbf{Covariate balance:} compute SMDs before and after matching using the pre-matching pooled SD \\
		\textbf{Output:} matched pairs $\mathcal{P}$
	\end{algorithm*}
	
	\begin{algorithm*}[p]
		\caption{Uplift modelling and evaluation (per cohort). Splits are formed at the pair level while ranking and evaluation are at the patient level.}
		\label{alg:uplift}
		\DontPrintSemicolon
		\LinesNotNumbered
		\textbf{Input:} matched pairs $\mathcal{P}$\\
		Split $\mathcal{P}$ 80/20 into train and test pairs\\
		Split the train pairs into 5 folds\\
		\textbf{Learners from causalml:} \texttt{BaseSClassifier}, \texttt{BaseTClassifier}, \texttt{BaseXClassifier}, \texttt{BaseRClassifier} and \texttt{UpliftRandomForestClassifier}\\
		\For{each learner $m$ and each candidate configuration}{
			\For{each fold $k$}{
				Fit $m$ on the other four folds and predict $\hat{\tau}(x)$ for the patients in fold $k$\\
				Rank fold $k$ patients by $\hat{\tau}(x)$ and compute AUQC with \texttt{qini\_score}
			}
			Record the mean and SD of AUQC across folds
		}
		Keep the best configuration per learner and select the learner $m^*$ with the highest mean AUQC\\
		Refit each learner on all train pairs and predict $\hat{\tau}(x)$ for test patients\\
		Compute test AUQC and Qini curves for all learners\\
		\textbf{Handcrafted rule with thresholds derived from train bin plots:} score $=$ BP grade (5-year window, 0 to 4), plus 1 if no prior hypertension diagnosis, plus 1 if 1 to 10 prescriptions (uncontrolled cohort only); and compute test AUQC\\
		Rank test patients by $\hat{\tau}_{m^*}(x)$ and form groups of the top 10\%, next 40\%, next 40\% and bottom 10\%\\
		\For{each group $g$}{
			Realised uplift $\gets \bar{Y}^{T=1}_{g} - \bar{Y}^{T=0}_{g}$ with a 95\% CI from the two-proportion standard error\\
			Mean predicted uplift $\gets$ mean of $\hat{\tau}_{m^*}(x)$ in $g$
		}
		\textbf{Feature importance:} \texttt{get\_importance} (permutation, on $\hat{\tau}(x)$ from the train set) for the S-learner; \texttt{feature\_importances\_} (mean decrease in impurity) for the uplift forest\\
		\textbf{Output:} AUQC table, Qini curves, group-level realised and predicted uplift, feature importances
	\end{algorithm*}

	\begin{table*}[htbp]
		\centering
		\caption{Base learners and hyperparameters for the five uplift models, selected separately for the full and uncontrolled BP cohorts via cross-validation. }
		\label{tab:hyperparams}
		\small
		\begin{tabular}{@{}p{0.17\linewidth}p{0.38\linewidth}p{0.38\linewidth}@{}}
			\toprule
			\textbf{Model} & \textbf{Full cohort} & \textbf{Uncontrolled BP cohort} \\
			\midrule
			S-learner &
			Logistic regression ($C=0.5$) &
			Random forest ($n_{\text{estimators}}=100$, $\text{max\_depth}=\text{None}$, $\text{min\_samples\_split}=0.1$, $\text{min\_samples\_leaf}=0.05$) \\
			\addlinespace
			T-learner &
			Logistic regression ($C=0.01$) &
			Logistic regression ($C=0.1$) \\
			\addlinespace
			X-learner &
			Outcome: SGD classifier (penalty=$\ell_2$, loss=log loss, $\alpha=0.1$) \newline
			Effect: linear regression &
			Outcome: SGD classifier (penalty=$\ell_2$, loss=log loss, $\alpha=0.1$) \newline
			Effect: elastic net ($\alpha=0.01$, $\text{l1\_ratio}=0.5$) \\
			\addlinespace
			R-learner &
			Outcome: SGD classifier (penalty=$\ell_2$, loss=log loss, $\alpha=0.1$, random\_state=seed) \newline
			Effect: linear regression &
			Outcome: SGD classifier (penalty=$\ell_2$, loss=log loss, $\alpha=0.1$, random\_state=seed) \newline
			Effect: elastic net ($\alpha=0.1$, $\text{l1\_ratio}=0.5$, random\_state=seed) \\
			\addlinespace
			Uplift random forest &
			$n_{\text{estimators}}=100$, $\text{max\_features}=10$, $\text{min\_samples\_leaf}=100$, $\text{max\_depth}=4$, $\text{random\_state}=111$ &
			$n_{\text{estimators}}=100$, $\text{max\_features}=10$, $\text{min\_samples\_leaf}=200$, $\text{max\_depth}=10$, $\text{random\_state}=111$ \\
			\bottomrule
		\end{tabular}
	\end{table*}

\begin{figure}[htbp]
\floatconts
{fig:calibration}
{\caption{Realised uplift (circles, with 95\% confidence intervals) and mean predicted uplift (diamonds) by predicted-uplift group on the held-out test set. Groups are the top 10\%, next 40\%, next 40\% and bottom 10\% by predicted uplift from the selected model in each cohort (uplift random forest for the full cohort and S-learner for the uncontrolled BP cohort).}}
{%
\subfigure[Full cohort]{
	\label{fig:calibration_full}
	\includegraphics[width=\linewidth]{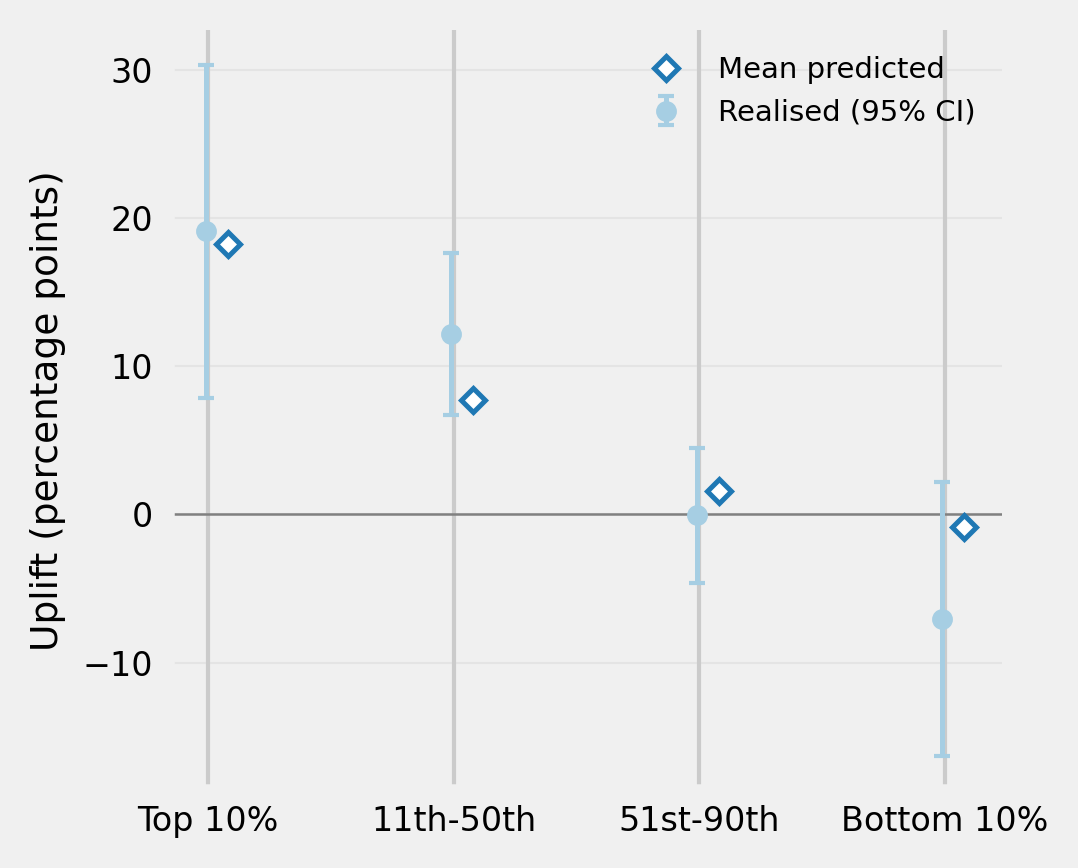}}%
\qquad
\subfigure[Uncontrolled BP cohort]{
	\label{fig:calibration_uncontrolled}
	\includegraphics[width=\linewidth]{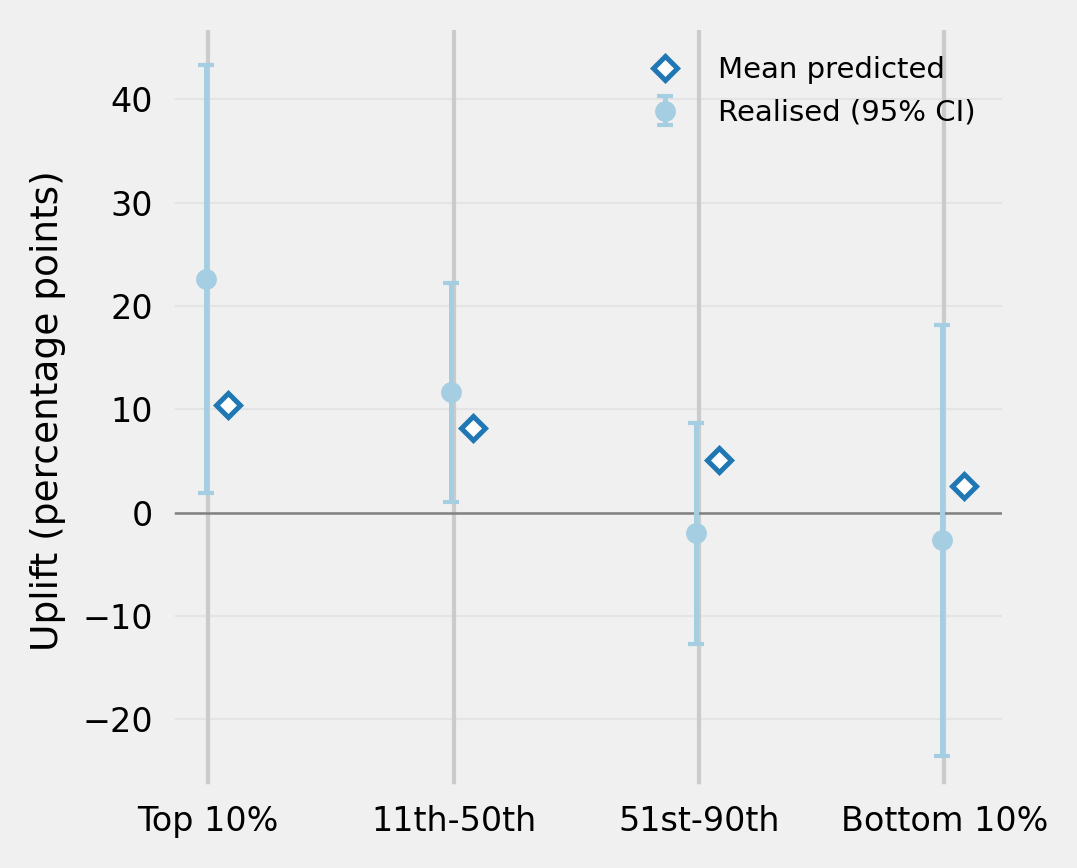}}
}
\end{figure}

\begin{table}[h!]
	\centering
	\caption{Covariate balance (standardised mean differences) within top-decile predicted uplift subgroup}
	\label{tab:topdecile_smd}
	\begin{tabular}{lrr}
	\toprule
	& \multicolumn{2}{c}{\textbf{SMD}} \\
	\cmidrule(l){2-3}
	\textbf{Covariate} & \textbf{Train} & \textbf{Test} \\
		\midrule
		Male & 0.050 & $-$0.078 \\
		Age & $-$0.231 & 0.069 \\
		Years since diagnosis & $-$0.395 & $-$0.056 \\
		BP grade (3-month) & $-$0.139 & 0.053 \\
		BP grade (1-year) & $-$0.113 & 0.089 \\
		BP grade (5-year) & $-$0.041 & 0.068 \\
		Household size & 0.025 & $-$0.053 \\
		Total CCI score & $-$0.111 & 0.066 \\
		Total diagnoses & $-$0.343 & $-$0.011 \\
		Total prescriptions & $-$0.353 & $-$0.041 \\
		\bottomrule
	\end{tabular}
\end{table}

\begin{table}[t]
	\centering
	\caption{Spearman correlation matrix of the disease duration and hypertension burden features.}
	\label{tab:corr}
	\begin{tabular}{@{}p{0.55\columnwidth}ccc@{}}
		\toprule
		& (1) & (2) & (3) \\
		\midrule
		(1) Years since diagnosis & 1.00 & & \\
		(2) Total diagnoses & 0.82 & 1.00 & \\
		(3) Total prescriptions & 0.74 & 0.79 & 1.00 \\
		\bottomrule
	\end{tabular}
\end{table}

\begin{figure*}[htbp]
	\floatconts
	{fig:featgroup}
	{\caption{Rate of BP control in the PTEC-HT and control groups within bins of each top feature for the full and uncontrolled BP cohort. The BP grades are N: normal BP, HN: high-normal BP, G1-G3: grades 1, 2, 3 hypertension.}}
	{%
		\subfigure[Full cohort]{
			\label{fig:featgroup_full}
			\includegraphics[width=0.9\linewidth,height=0.4\textheight,keepaspectratio]{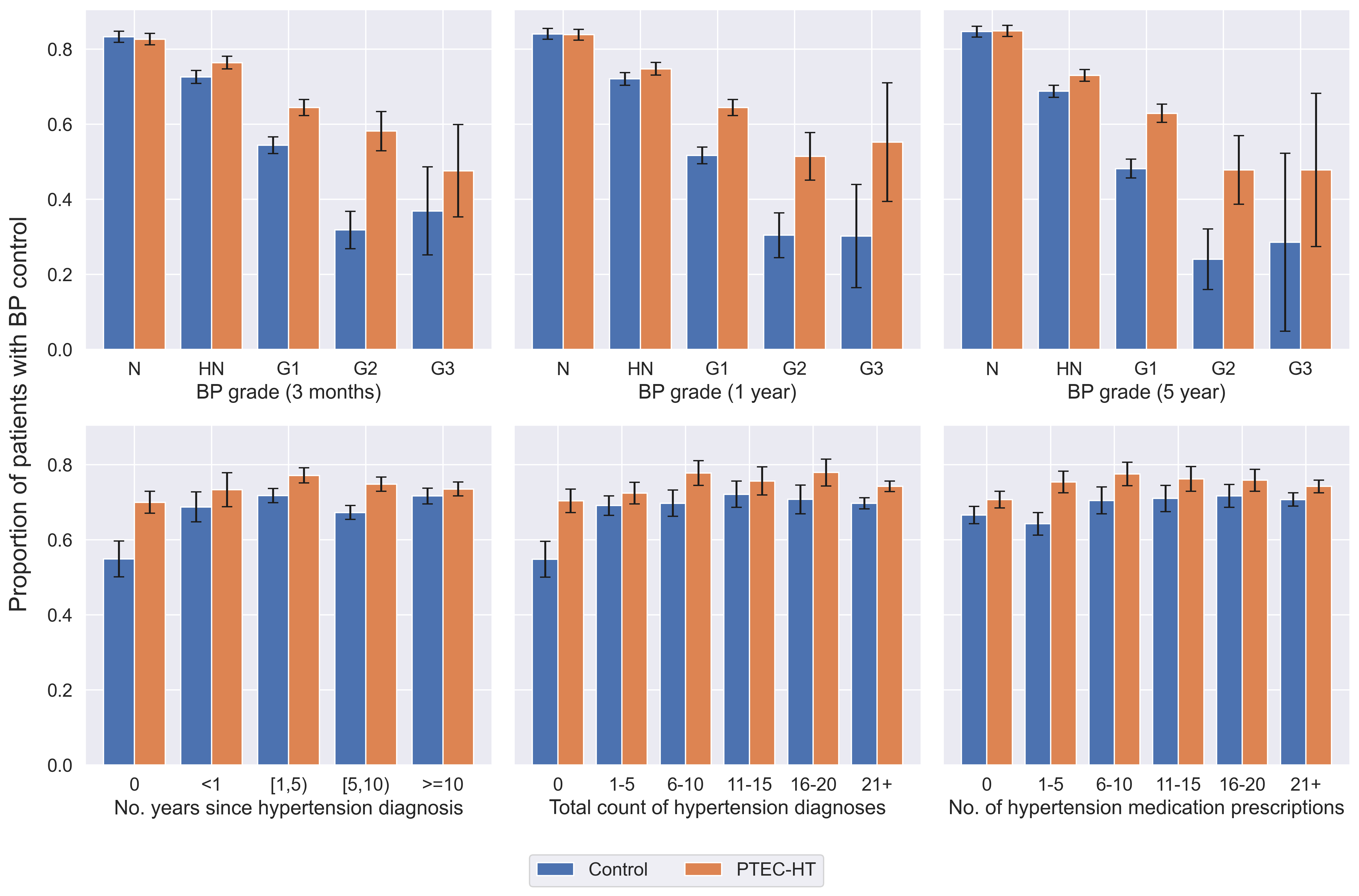}}%
		\qquad
		\subfigure[Uncontrolled BP cohort]{
			\label{fig:featgroup_uncontrolled}
			\includegraphics[width=0.9\linewidth,height=0.4\textheight,keepaspectratio]{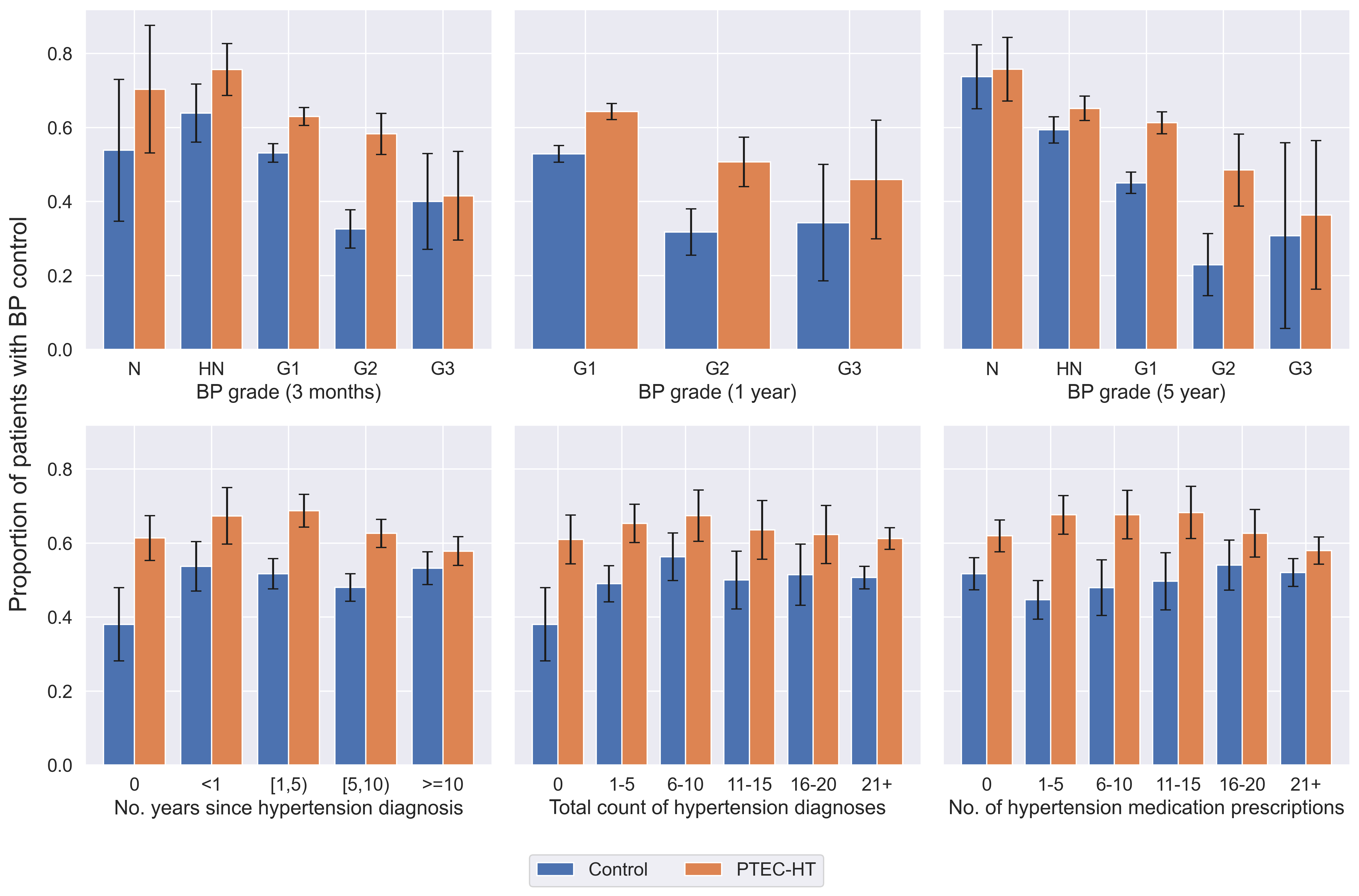}}%
	}
\end{figure*}

\end{document}